\documentclass{article}

\usepackage[final,nonatbib]{neurips_2025}
\usepackage[numbers]{natbib}

\usepackage[utf8]{inputenc} %
\usepackage[T1]{fontenc}    %
\usepackage{hyperref}       %
\usepackage{url}            %
\usepackage{booktabs}       %
\usepackage{amsfonts}       %
\usepackage{nicefrac}       %
\usepackage{microtype}      %
\usepackage{xcolor}         %
\usepackage{enumitem}
\usepackage{graphicx}
\usepackage{amsmath}
\usepackage{booktabs}
\usepackage{multirow}
\usepackage{colortbl}
\usepackage{makecell}
\usepackage{diagbox}
\usepackage{cleveref}
\usepackage{media9}
\usepackage{animate}
\usepackage{adjustbox}
\usepackage{capt-of}
\usepackage{subcaption}

\title{4DGT: Learning a 4D Gaussian Transformer \\ Using Real-World Monocular Videos}

\author{%
\textbf{Zhen Xu}\textsuperscript{1,2,*}
\textbf{Zhengqin Li}\textsuperscript{1}
\textbf{Zhao Dong}\textsuperscript{1}
\textbf{Xiaowei Zhou}\textsuperscript{2}
\textbf{Richard Newcombe}\textsuperscript{1}
\textbf{Zhaoyang Lv}\textsuperscript{1}\\
{\small \textsuperscript{1}Reality Labs Research, Meta} \quad \textsuperscript{2}Zhejiang University\\
{\small \textit{Project page:} \url{https://4dgt.github.io}}
}

\usepackage{xcolor}
\definecolor{myred}{rgb}{0.8,0,0}
\definecolor{mygreen}{rgb}{0,0.8,0}
\definecolor{myblue}{rgb}{0,0,0.95}
\definecolor{mypurple}{rgb}{0.75,0,0.75}
\definecolor{myorange}{rgb}{0.75,0.25,0.25}
\usepackage{pifont}

\newcommand{\nbf}[1]{{\noindent \textbf{#1.}}}

\newcommand{\stddev}[1]{\footnotesize$^{\pm#1}$}
\newcommand{\second}[1]{\cellcolor{orange!25}#1}
\newcommand{\best}[1]{\cellcolor{red!25}#1}

\newcommand{\camrdy}[1]{#1}

\begin{document}

\maketitle

\begin{center}
    \centering
    \captionsetup{type=figure}
    \includegraphics[width=1.0\textwidth]{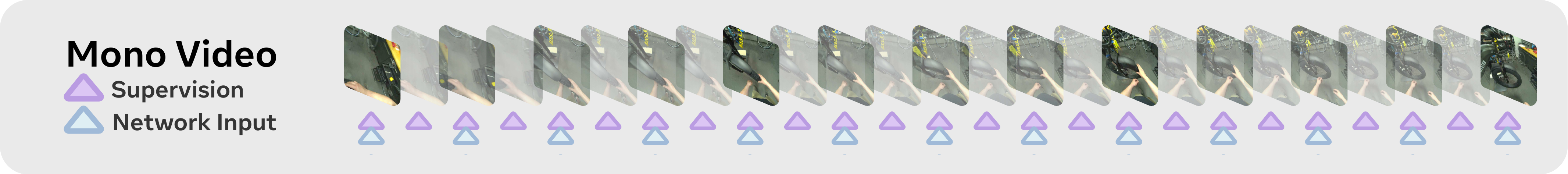}
    \includegraphics[width=0.1375\textwidth]{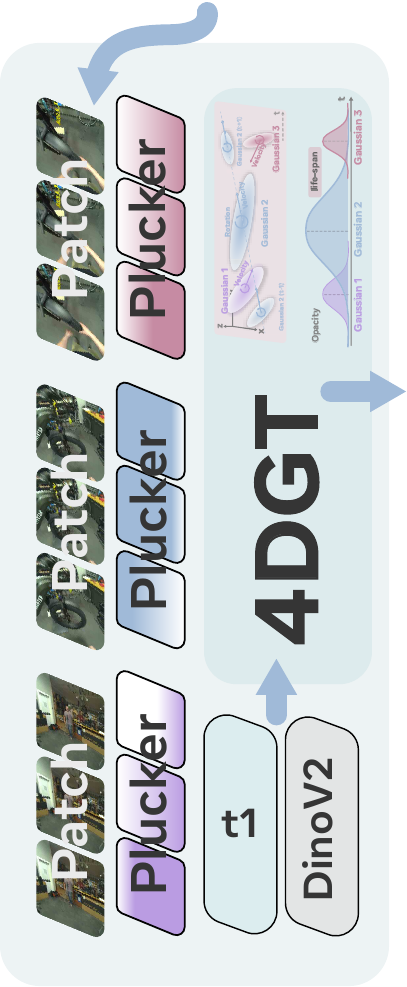}
    \includegraphics[width=0.3725\textwidth]{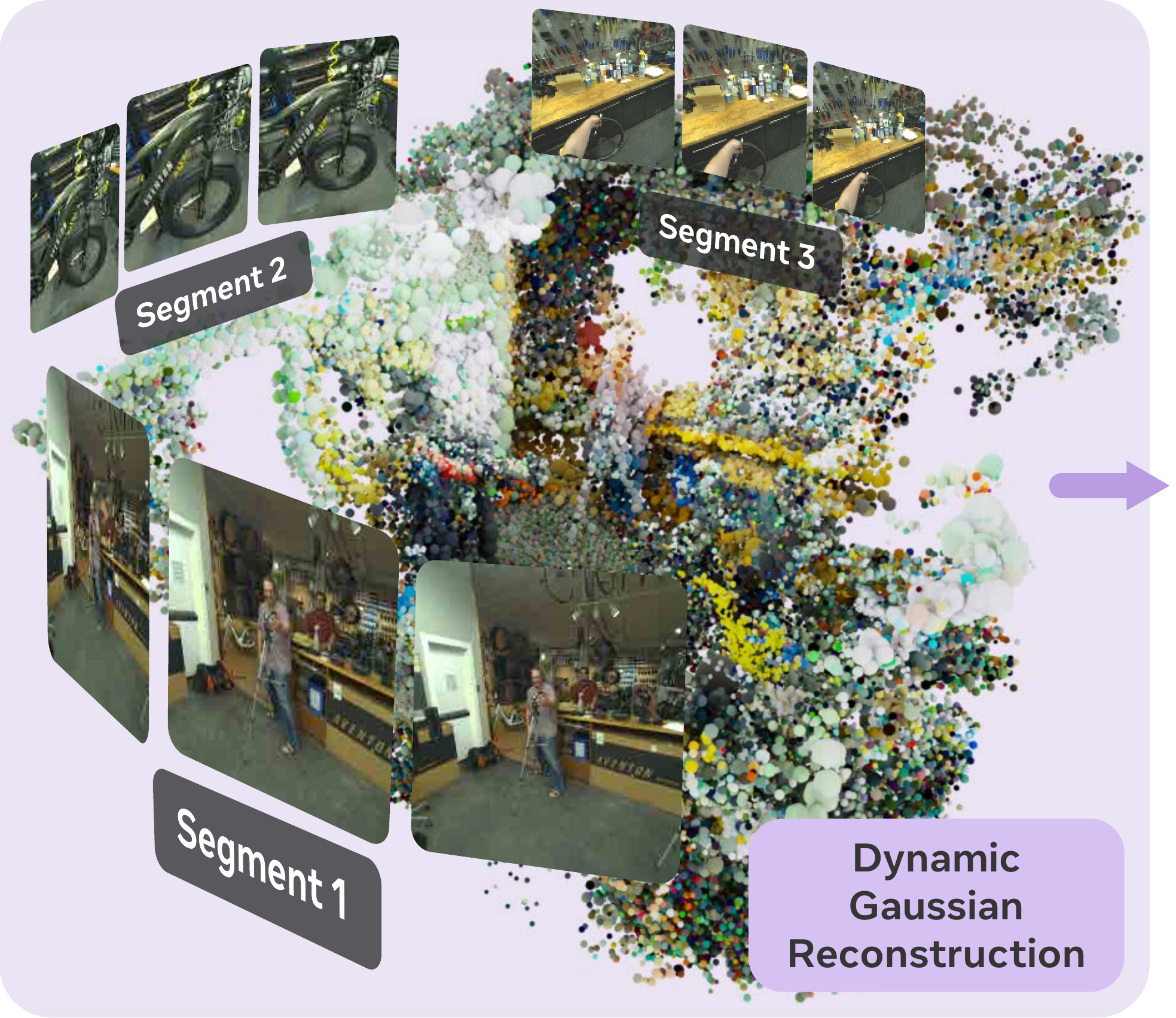}
    \animategraphics[autoplay,loop,every=1,width=0.4775\textwidth,poster=0]{10}{videos/teaser/img_}{042}{127}
    \captionof{figure}{
        We propose a scalable 4D dynamic reconstruction model trained only on real-world monocular RGB videos.
        The feed-forward 4DGS (\cref{sec:4dgs}) representation enables us to render the geometry and appearance of the dynamic scene from novel views in real-time.
        Even without explicit supervision, the model can learn to distinguish dynamic contents from the background and produce realistic optical flows.
        The figure shows an enlarged set of Gaussians for the purpose of visualization.
        \textbf{The embedded rendered videos only play in Adobe Reader or KDE Okular.}
    }
    \label{fig:teaser}
    \vspace{1em}
\end{center}

\begin{abstract}
    We propose 4DGT, a 4D Gaussian-based Transformer model for dynamic scene reconstruction, trained entirely on real-world monocular posed videos. Using 4D Gaussian as an inductive bias, 4DGT unifies static and dynamic components, enabling the modeling of complex, time-varying environments with varying object lifespans. We proposed a novel density control strategy in training, which enables our 4DGT to handle longer space-time input and remain efficient rendering at runtime. Our model processes 64 consecutive posed frames in a rolling-window fashion, predicting consistent 4D Gaussians in the scene. Unlike optimization-based methods, 4DGT performs purely feed-forward inference, reducing reconstruction time from hours to seconds and scaling effectively to long video sequences.
    Trained only on large-scale monocular posed video datasets, 4DGT can outperform prior Gaussian-based networks significantly in real-world videos and achieve on-par accuracy with optimization-based methods on cross-domain videos.

\end{abstract}

\renewcommand{\thefootnote}{\roman{footnote}}
\footnotemark[0]
\footnotetext[0]{* Work done during internship at Meta.}

\section{Introduction}
\label{sec:introduction}

Humans record videos to digitize interactions with their surroundings. The ability to recover persistent geometry and 4D motion from videos has a profound impact on AR/VR, robotics, and content creation. Modeling 4D dynamic interactions from general-purpose videos remains a long-standing challenge. Prior work relying on multi-view synchronized capture or depth sensing is constrained to specific application domains. Recent progress in monocular dynamic video reconstruction via per-video optimization shows promise, but lacks scalability due to its time-consuming inference.

In this paper, we propose 4D Gaussian Transformer (4DGT), a novel transformer-based model that reconstructs dynamic scenes from posed monocular videos in a feedforward manner. We assume camera calibration and 6-degree-of-freedom (6DoF) poses are available from on-device SLAM \cite{engel2023project} or offline pipelines \cite{li2024megasam,murai2024mast3rslam}. Inspired by recent feedforward reconstruction methods for static 3D scenes \cite{gslrm2024, ziwen2024long}, 4DGT learns reconstruction from data and adopts 4D Gaussian Splatting (4DGS) \cite{wang2024freetimegs} as a unified representation for both static and dynamic content, differing only in lifespan. This design enables fast 4D reconstruction from short videos in seconds. For longer videos with global pose consistency, 4DGT predicts consistent world-aligned 4DGS using 64-frame rolling windows.

Training a 4D representation is challenging in defining appropriate supervision. While dynamic monocular videos are abundant, they lack space-time constraints. Multi-view video datasets \cite{broxton2020immersive,li2022neural} are limited in both quantity and diversity, making them insufficient for training models that generalize in the wild. Prior methods~\cite{ren2024l4gm} trained on synthetic object-level data suffer from a generalization gap when applied to complex real-world dynamics.

To address this, we train 4DGT exclusively on posed monocular videos from public datasets. We use two key strategies to mitigate space-time ambiguity. First, we leverage depth and normal predictions from expert models \cite{piccinelli2025unidepthv2, yang2024depth, ye2024stablenormal} as auxiliary supervision for guiding geometry learning. Second, we regularize predicted Gaussian properties to favor longer lifespans and reduce overfitting to specific views. These enable 4DGT to effectively disentangle space-time structure, yielding high-quality geometry, novel view synthesis, and emergent motion properties such as segmentation and flow. Our reconstructions also show better metric consistency than the expert models used for supervision.

Scaling a transformer to predict dense, pixel-aligned 4DGS presents two main challenges. First, dense pixel-aligned 4DGS predictions are computationally expensive for training and rendering. Inspired by density control in 3DGS \cite{kerbl20233d}, we introduce a pruning strategy that removes the redundant pixel-aligned 4DGS and further increases tokens in a second training with denser space-time samples. This effectively reduces 80\% of Gaussians and enables a $16\times$ higher sampling rate with the same compute. Second, as space-time samples increase, the number of tokens grows, and vanilla self-attention scales quadratically. To address this, we propose a level-of-detail structure via multi-level spatiotemporal attention, achieving an additional $2\times$ reduction in computational cost.

4DGT is the first transformer-based method for predicting 4DGS in a feedforward manner using only real-world posed monocular videos in training. Extensive evaluations across datasets and domains show that 4DGT achieves comparable reconstruction quality to optimization-based methods while being three orders of magnitude faster, making it practical for long video reconstruction. \camrdy{Compared to prior methods that only train on synthetic object-level data~\cite{ren2024l4gm}, 4DGT generalizes better to complex real-world dynamics. Compared to the per-frame prediction pipeline~\cite{liang2024feed}, it also exhibits emergent motion properties.}

In summary, we make the following technical contributions:
\begin{itemize}[leftmargin=*]
    \setlength\itemsep{0em}
    \item We introduce 4DGT, a novel 4DGS transformer trained on posed monocular videos at scale, which produces consistent 4D video reconstructions in seconds at inference.
    \item We propose a training strategy to densify and prune space-time pixel-aligned Gaussians, reducing 80\% of predictions, achieving 16× higher sampling rate during training and a 5× speed-up in rendering.
    \item We design a multi-level attention module to efficiently fuse space-time tokens, further reducing training time by half.
    \item Our experiments demonstrate strong scalability of 4DGT across real-world domains using mixed training datasets and can outperform the previous Gaussian network significantly. The performance of 4DGT is on par with optimization-based methods in accuracy in cross-domain videos recorded by similar devices used in training, while being 3 orders of magnitude faster.
\end{itemize}

\section{Related Work}
\label{sec:related_work}

\nbf{Nonrigid reconstruction} Recovering dynamic content from video has long been a holy grail challenge in 3D vision. Early approaches demonstrated promising non-rigid shape reconstruction from RGB-D videos \cite{newcombe2015dynamicfusion, slavcheva2017killingfusion, bozic2020deepdeform}, but relied heavily on depth input and struggled with complex dynamic scenes. Since the seminal work on Neural Radiance Fields (NeRF) \cite{mildenhall2020nerf}, several methods have extended NeRF to 4D using multi-view videos \cite{li2022neural,fridovich2023k} or posed monocular videos \cite{park2021nerfies, park2021hypernerf}. However, 4D NeRFs are slow to train and render, limiting their scalability for complex dynamic scenes. Recently, generative priors have shown promise in aiding 4D reconstruction \cite{wu2024cat4d}, offering strong regularization for shape and motion across space and time. Still, optimizing 4D representations remains time-consuming, and reconstruction quality depends heavily on the generalizability of priors across different scene domains.

\nbf{Dynamic Gaussian representations} Since the introduction of 3D Gaussian Splatting \cite{kerbl20233d}, several methods have extended it to 4DGS variants \cite{yang2023gs4d,xu2024longvolcap,wang2024freetimegs,duan20244d}, showing promising dynamic scene reconstruction from multi-view videos with faster training and real-time rendering support. However, optimizing dynamic Gaussians for monocular videos remains challenging. Recently, a few works have shown that complex 4D scenes composed of moving Gaussians can be recovered by leveraging depth, segmentation, and tracking priors from 2D expert models \cite{wu20234d, wang2024shape, lei2024mosca}. Despite strong performance, these methods involve complex processing, including manual annotation on dynamic regions. It further requires lengthy optimization, limiting its scalability in practical applications.

\nbf{Large reconstruction models} Transformer-based 3D large reconstruction models (LRMs) have shown strong potential for learning high-quality 3D reconstruction from data, at the object level \cite{hong2023lrm,jiang2023consistent4d,pan2024efficient4d,xie2024sv4d,li2025lirm} and static scenes \cite{xie2024lrm,gslrm2024,ziwen2024long,xu2024grm}. LRMs can generate reconstructions in seconds from a few input views, achieving quality comparable to optimization-based neural methods. However, training LRMs requires large-scale multi-view supervision of the same instance, which is readily available in synthetic datasets or static scene captures, but remains scarce for real-world videos.

Some recent efforts explored training transformers to predict time-dependent 3D-GS \camrdy{\cite{ren2024l4gm,liang2024feed,yang2025storm,qi2025predicting}} using animated synthetic data \cite{ren2024l4gm}, self-curated real-world internet videos \cite{liang2024feed} and street-level data \cite{yang2025storm}, making them the closest related works in motivation. In contrast to these methods, which predict time-dependent 3D-GS representations, our 4DGT offers a holistic 4D scene representation that captures geometry better and enables motion understanding capabilities lacking in prior approaches. \cite{yang2025storm} requires multi-camera input and only focuses on street-level scenes. Compared to training on synthetic data in \cite{ren2024l4gm}, our real-world training approach generalizes better to real-world scenes. \camrdy{Compared to B-Timer \cite{liang2024feed}, which adopts a per-frame Gaussian prediction pipeline, our method produces explicit dynamic Gaussians thus can model explicit motion, showing emergent capabilities like motion segmentation. Compared to Pred. 3D Repr. \cite{qi2025predicting}, which adopts a tri-plane based implicit representation, our Gaussian model enables fast rendering after the feed-forward reconstructrion.}

\section{Method}
\label{sec:method}
\begin{figure}[t]
    \centering
    \includegraphics[width=1.0\textwidth]{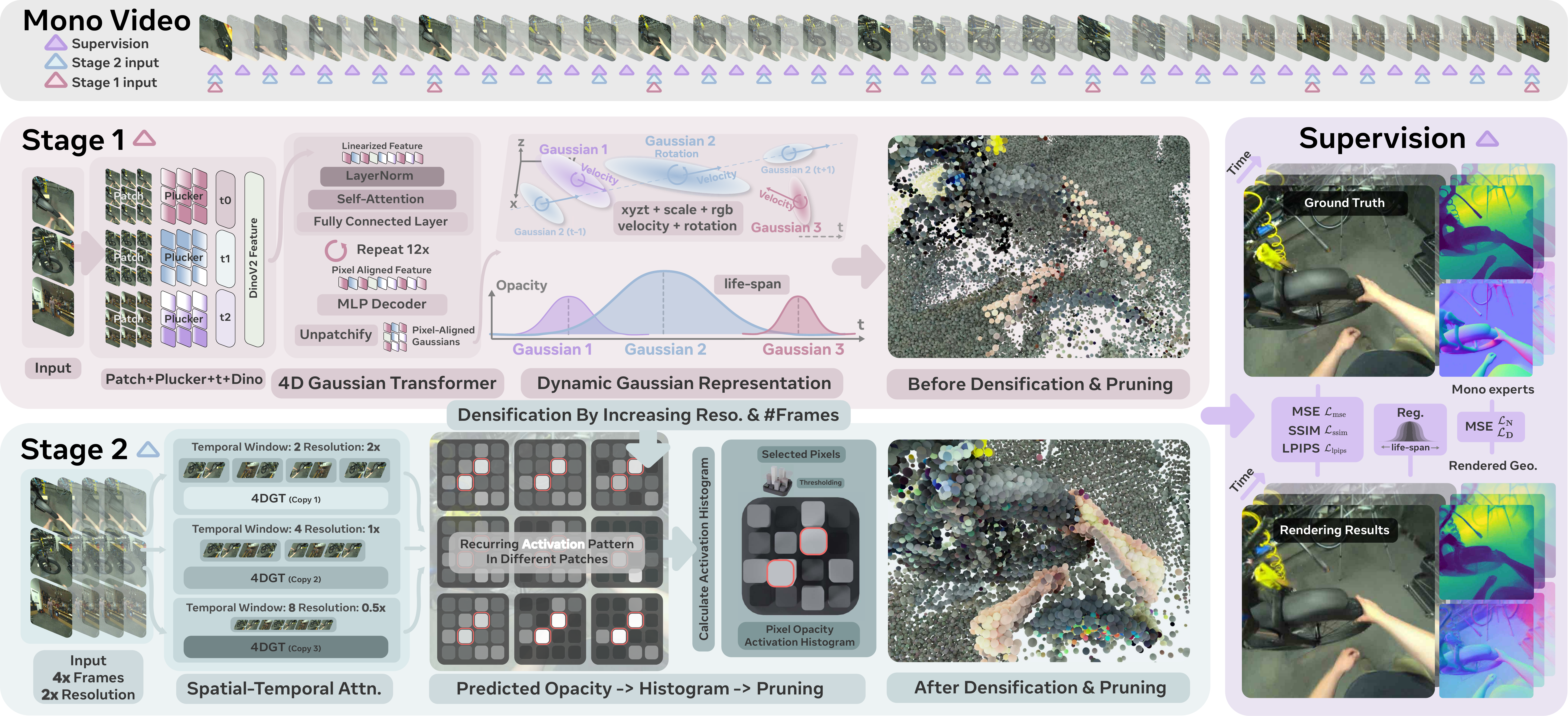}
    \caption{An overview of our method in training and rendering. 4DGT takes a series of monocular frames with poses as input. During training, we subsample the temporal frames at different granularity and use all images in training. We first train 4DGT to predict pixel-aligned Gaussians at coarse resolution in stage one. In stage two training, we pruned a majority of non-activated Gaussians according to the histograms of per-patch activation channels, and densify the Gaussian prediction by increasing the input token samples in both space and time. At inference time, we run the 4DGT network trained after stage two. It can support dense video frames input at high resolution.}
\end{figure}

Given a posed monocular video, 4DGT uses a transformer to predict a set of 4DGS, which can be rendered in real time. We first describe the architecture and dynamic scene representation in \cref{sec:4dgs}. To enable efficient training and rendering at scale, we introduce a pixel density control strategy and a level-of-detail structure based on spatial attention. Both techniques improve space-time sampling rates under fixed compute budgets. In \cref{sec:training}, we detail our training process and regularization strategies designed to resolve space-time ambiguities in monocular videos.

\subsection{Feed-Forward Dynamic Gaussian Prediction}
\label{sec:4dgs}

\nbf{Input encoding}
Given a posed monocular video, we extract a set of image frames $\mathbf{I}_i$ with camera calibration $\mathbf{P}_i$ and timestamp $\mathbf{T}_i$, denoted as $ \{\mathbf{I}_i \in R^{H \times W \times 3}, \mathbf{P}_i \in R^{H \times W \times 6}, \mathbf{T}_i \in R^{H \times W \times 1}| i = 1 \cdots N\}$, where $\mathbf{P}_i$ represents the Plücker coordinates \cite{julius1865new} and $N$ is the total number of frames.
We convert them into patches.
For frame $i$, the patches are denoted as $\{\mathbf{I}_{i,j} \in R^{p \times p \times 3} | j = 1 \cdots HW/p^2\}$, $\{\mathbf{T}_{i,j}\}$ and $\{\mathbf{P}_{i,j}\}$, where $p$ is the patch size.

\nbf{Feature fusion} We use the pretrained DINOv2 image encoder \cite{oquab2023dinov2} to extract high-level $C$-dimensional features $\mathbf{F}_{i,j} \in R^{C}$.
These are concatenated with the temporal and spatial encoding $\mathbf{T}_{i,j}$ and $\mathbf{P}_{i,j}$ as well as the input RGB image $\mathbf{I}_{i,j}$ to form the fused transformer input:
\begin{equation}
  \{\mathbf{X}_{i,j} \} = \mathcal{F}(\{\mathbf{I}_{i,j} \oplus \mathbf{T}_{i,j} \oplus \mathbf{P}_{i,j} \oplus \mathbf{F}_{i,j}\ | i = 1 \cdots N, j = 1 \cdots HW/p^2\}),
\end{equation}
where $\oplus$ denotes the concatenation operation and $\mathcal{F}$ denotes the all-to-all self-attention transformer module.
In contrast to ViT, input used static LRMs in that it uses only Plücker rays \cite{gslrm2024} or DINO feature \cite{hong2023lrm}, our transformer takes timestamp-aware Plücker rays with DINO feature together as input, which we found to be beneficial to provide the best prediction in view synthesis as well as geometry prediction.

\nbf{Dynamic Gaussians}
\camrdy{We use a variant of 4DGS \cite{yang2023gs4d,wang2024freetimegs} to unify the various components in dynamic scene predictions.}
To better represent geometry, we adopt the 2DGS \cite{huang20242d} defined by the center $\mathbf{x} \in R^3$, scale $\mathbf{s} \in R^2$, opacity $\mathbf{o} \in R^1$ and orientation $\mathbf{q} \in R^4$ (quaternion) of the Gaussians. Compared to 3DGS\cite{kerbl20233d} used in previous work \cite{yang2023gs4d,wang2024freetimegs}, 2DGS yields better geometry predictions.
To represent motion, we use four temporal attributes, namely the temporal center $\mathbf{c} \in R^1$, life-span $\mathbf{l} \in R^1$, velocity $\mathbf{v} \in R^3$, and angular velocity $\boldsymbol{\omega} \in R^3$ (as angle-axis) for each Gaussian. %
Given a specific timestamp $t_{s}$ for rendering a particular dynamic Gaussian point $\mathbf{g} = \{\mathbf{x}, \mathbf{s}, \mathbf{q}, \mathbf{o}, \mathbf{c}, \mathbf{l}, \mathbf{v}, \boldsymbol{\omega}\}$, we first calculate the offset to the opacity, location and orientation of the Gaussian point from the temporal attributes \cite{yang2023gs4d}.
Specifically, the life-span $\mathbf{l}$ is used to influence the Gaussian opacity $\mathbf{o}$ over time:

\begin{align}
  \boldsymbol{\sigma} = \sqrt{-\frac{1}{2} \cdot \frac{({\mathbf{l}}/{2})^2}{\log(o_{th})}},\ \
  \mathbf{o}_{t_{s}} = \mathbf{o} \cdot e^{-\frac{1}{2} \cdot \frac{(t_{s} - \mathbf{c})^2}{\boldsymbol{\sigma}^2}},
\end{align}

where $o_{th}$ is the opacity multiplier at the life-span boundary and $\boldsymbol{\sigma}$ is the standard deviation of the Gaussian distribution in the temporal domain.
Intuitively, the Gaussian retains its full opacity at its temporal center and fades in a Gaussian distribution along the temporal axis.
At ${\mathbf{l}}/{2}$ time relative to the temporal center $\mathbf{c}$, the opacity of the point is reduced by multiplying a small factor $o_{th}$, which is set to $0.05$ for all experiments.
The location and orientation of each Gaussian is adjusted by the velocity $\mathbf{v}$ and angular velocity $\boldsymbol{\omega}$ to account for the motion:
\begin{align}
  \mathbf{x}_{t_{s}} = \mathbf{x} + \mathbf{v} \cdot (t_{s} - \mathbf{c}),\ \
  \mathbf{q}_{t_{s}} = \mathbf{q} \cdot \phi({\boldsymbol{\omega} \cdot (t_{s} - \mathbf{c})}),
\end{align}

where $\phi$ denotes converting the angle-axis representation to a quaternion.
For each 2DGS with $\mathbf{x}_{t_{s}}, \mathbf{s}_{t_{s}}, \mathbf{q}_{t_{s}}, \mathbf{o}_{t_{s}}$ at timestamp $t_{s}$, we apply 2DGS rasterizer implemented in \cite{ye2025gsplat} to render image.

\nbf{Dynamic Gaussian decoding}
Given the pixel-aligned features $\{\mathbf{X}_{i,j}\}$, we use a transformer to decode the pixel-aligned 4DGS for each frame as
\begin{equation}
  \label{eq:gs_prediction}
  \{\mathbf{G}_{i,j} \} = \mathcal{D}_{\mathbf{x},\mathbf{s},\mathbf{q},\mathbf{o},\mathbf{c},\mathbf{l},\mathbf{v},\boldsymbol{\omega}}(\{\mathbf{X}_{i,j} | i = 1 \cdots N, j = 1 \cdots HW/p^2\}),
\end{equation}
where $\mathcal{D}_{\mathbf{x},\mathbf{s},\mathbf{q},\mathbf{o},\mathbf{c},\mathbf{l},\mathbf{v},\boldsymbol{\omega}}$ denotes the MLP-based decoder head for producing the full suite of dynamic Gaussian parameters $\{\mathbf{G}_{i,j} \}$.

Our proposed 4DGS can unify the prediction of appearance and geometry properties of both static and dynamic elements.
For static scenes, the network can learn to predict Gaussians with a long-living lifespan $\mathbf{l} \rightarrow \infty, \mathbf{v} \rightarrow 0, \boldsymbol{\omega} \rightarrow 0$. For complex dynamic motions with occlusions, it predicts transient dynamic objects with short-living Gaussians $\mathbf{l} \rightarrow 0$.

\subsection{Multi-level Pixel \& Token Density Control}
\label{sec:density_control}

While pixel-aligned Gaussian has been a standard choice in prior work \cite{gslrm2024,ziwen2024long}, it has a severe limitation in representing video frames that require dense, long-term sampling to capture motion effectively. Naively sampling frames spatial-temporally would result in two key issues. First, the increasing number of aligned Gaussians degrades optimization and rendering performance, leading to suboptimal training and blurry details in dynamic regions. Second, the growing number of input tokens significantly increases computational cost, resulting in under-trained models.

\nbf{Two-stage training} We introduce a two-stage approach to address these challenges. First, we train on coarsely sampled low-resolution images from scratch until convergence. In the second stage, inspired by 3DGS \cite{kerbl20233d}, we propose to filter pixel-aligned Gaussians by pruning low-opacity predictions per patch based on the histograms and increasing the token count to predict more Gaussians across space-time. Additionally, we introduce a multi-level spatiotemporal attention mechanism to further reduce the computational cost of self-attention layers.

\nbf{Pruning} %
After the initial training stage at coarse resolution, we compute a histogram of activated Gaussians per patch and observe that only a few channels are activated. A similar pattern emerges in other pixel-aligned Gaussian methods \cite{gslrm2024},
motivating us to decode only a small set of Gaussians using the activated channels.
Formally, for each patch of Gaussian parameters $\mathbf{G}_{i,j}$, we consider the standard deviation of their opacity values $ \mathbf{o}_{i,j} = \{ \mathbf{o}_{i,j,k} | k = 1 \cdots p^2 \}$:
\begin{align}
  \mu(\mathbf{o}_{i,j}) = \frac{1}{p^2} \sum_{k=1}^{p^2} \mathbf{o}_{i,j,k}, \ \
  \sigma(\mathbf{o}_{i,j}) = \sqrt{\mu(\mathbf{o}_{i,j}^{2}) - \mu(\mathbf{o}_{i,j})^2},
\end{align}
where $\mu(\mathbf{o}_{i,j})$ is the mean of the opacity values and $\sigma(\mathbf{o}_{i,j})$ is the standard deviation.
A particular pixel $k$ is considered activated if it has a value larger than 1 unit of the standard deviation:
\begin{align}
  \mathbf{m}_{i,j,k} = \begin{cases}
                         1, & \mathbf{o}_{i,j,k} > \mu(\mathbf{o}_{i,j}) + \sigma(\mathbf{o}_{i,j}), \\
                         0, & \text{otherwise}.
                       \end{cases}
\end{align}
where $k$ is the index of the pixel in the patch and $\mathbf{m}_{i,j,k}$ indicates whether the pixel is activated.
\camrdy{
  A histogram $\mathbf{h}_{i,j}$ of all activation mask $\mathbf{m}_{i,j} = \{ \mathbf{m}_{i,j,k} | k = 1 \cdots p^2 \}$ for the patch output is:
  \begin{align}
    \label{eq:histogram_calculation}
    \mathbf{H} = \sum_{i=1,\,j=1}^{N,\,HW/p^2} \mathbf{M}_{i,j},
  \end{align}
  where $\mathbf{M}_{i,j} \in \mathbb{N}^{p^2}$ is the activation mask for patch $(i,j)$, $\mathbf{H} \in \mathbb{N}^{p^2}$ is the aggregated histogram of all activation masks, and $p$ is the patch size.
}
We select $S$ channel from the histogram $\mathbf{h}_{i,j}$ for the patch output for all patches onward in training.
This effectively implements an $S / p^2$ times reduction in the number of Gaussians for each patch, mimicking the pruning strategy of 3DGS \cite{kerbl20233d}. We provide more in-depth analysis for this histogram-based pruning strategy compared to alternatives with visualizations in the appendix.

\nbf{Densification} The predicted Gaussian number can naturally increase with more space-time token inputs, either in resolution per frame or temporal frame numbers.
The initially trained model provides a good scaffolding for pixel-aligned Gaussians when we increase the input token number in space and time.
In the second stage of training, we increase the spatial and temporal resolution by a factor of $R_{s}$ and $R_{t}$, respectively.
Combining the densification process and pruning strategy, this would result in $R_{s}^2 \cdot R_{t} \cdot {S} / {p^2}$ times of the Gaussians compared to the first stage.
We select $R_{s} = 2$, $R_{t} = 4$, $S = 10$, and $p = 14$ for all experiments, leading to only $80\%$ of the original number of Gaussians while greatly increasing the sampling rate of space-time by $16$ times.

\nbf{Multi-level spatial-temporal attention}
\label{sec:temporal_lod}
The number of patches participating in the self-attention module $\mathcal{F}$ increases by a factor of $R_{s}^2 \cdot R_{t},$ which will slow down optimization and inference significantly.
To mitigate this, we propose a temporal level-of-detail attention mechanism to reduce the computational cost.
We propose to divide the $N$ input frames into $M$ equal trunks in the highest level.
This division limits the attention mechanism in the temporal dimension, but reduces the computation of calculating $n$ total tokens to $O(\frac{n^2}{M})$. To balance spatial-temporal samples, we construct a temporal level-of-detail structure by alternating the temporal range and spatial resolution, achieving a much smaller overhead while maintaining the ability to handle long temporal windows. For each level $l$, we reduce the spatial resolution by a factor of $2^l$ and increase temporal samples by 2.
Empirically, we use level $L=3$ and $M=4$, which leads to an approximately $2$ times reduction in the computational cost.

\subsection{Training}
\label{sec:training}

\nbf{Loss and regularization}
We train 4DGT using segments of $W=128$ consecutive frames from the monocular video and subsample every $8$ frames as input, resulting in $N=16$ input frames.
Notably, for the second stage training where we apply techniques mentioned in \cref{sec:density_control} and \cref{sec:temporal_lod}, we increase the number of input frames to $N=64$.
After obtaining all Gaussian parameters $\{\mathbf{G}_{i,j} \}$ from each of the $N$ input frames, we render them to all $W=128$ images for self-supervision and compute the MSE loss.
Additionally, we add the perceptual LPIPS loss \cite{johnson2016perceptual} $\mathcal{L}_{lpips}$ and SSIM loss \cite{wang2004image} $\mathcal{L}_{ssim}$ for better perceptual quality.
\begin{align}
  \label{eq:imag_rendering_loss}
  \mathcal{L}_{\mathrm{mse}} = \sum_{i=1}^{W} \frac{\left\| \mathbf{I}_{i} - \mathbf{I}_{i}^{'} \right\|_2}{W},\ \
  \mathcal{L}_{\mathrm{lpips}} = \sum_{i=1}^{W} \frac{\left\| \psi(\mathbf{I}_{i}) - \psi(\mathbf{I}_{i}^{'}) \right\|_{1}}{W},\ \
  \mathcal{L}_{\mathrm{ssim}} = \sum_{i=1}^{W} \frac{\mathrm{SSIM}(\mathbf{I}_{i}, \mathbf{I}_{i}^{'})}{W},
\end{align}
where $\mathbf{I}_{i}$ denotes the input image and $\mathbf{I}_{i}^{'}$ is the rendered image, $\psi$ is the pre-trained layers AlexNet \cite{krizhevsky2012imagenet} and $\mathrm{SSIM}$ is the SSIM function \cite{wang2004image}.
To better regularize the training, we encourage the points to be static and have a long lifespan using:
\begin{align}
  \mathcal{L}_{\mathbf{v}} = \sum_{i=1,j=1,k=1}^{N,HW/p^2,p^2} \frac{\left\| \mathbf{v}_{i,j,k} \right\|_1}{NHW},\ \
  \mathcal{L}_{\boldsymbol{\omega}} = \sum_{i=1,j=1,k=1}^{N,HW/p^2,p^2} \frac{\left\| \boldsymbol{\omega}_{i,j,k} \right\|_1}{NHW},\ \
  \mathcal{L}_{\mathbf{l}} = \sum_{i=1,j=1,k=1}^{N,HW/p^2,p^2} \frac{\left\| \frac{1}{\mathbf{l}_{i,j,k}} \right\|_1}{NHW},
  \label{eq:regularization_loss}
\end{align}
where $\mathbf{v}_{i,j,k}$ is the velocity of the Gaussian point, $\boldsymbol{\omega}_{i,j,k}$ is the angular velocity of the Gaussian point and $\mathbf{l}_{i,j,k}$ is the life-span of the Gaussian point.

\nbf{Expert guidance}
We observe that training can benefit from leveraging monocular export models in geometry prediction.
We extract the depth map $\mathbf{D}_{i}$ and normal map $\mathbf{N}_{i}$ from all $W$ frames using DepthAnythingV2 \cite{yang2024depth} and StableNormal \cite{ye2024stablenormal} and use them as a pseudo-supervision signal:
\begin{align}
  \mathcal{L}_{\mathbf{D}} = \sum_{i=1}^{W} \frac{ \left\| \mathbf{D}_{i} - \mathbf{D}_{i}^{'} \right\|_2}{W},\ \
  \mathcal{L}_{\mathbf{N}} = \sum_{i=1}^{W} \frac{ \left\| \mathbf{N}_{i} - \mathbf{N}_{i}^{'} \right\|_2}{W},
\end{align}
where $\mathbf{D}_{i}^{'}$ and $\mathbf{N}_{i}^{'}$ are the predicted depth and normal map rendered using the 2DGS rasterizer \cite{huang20242d}.
The final loss function for training the feed-forward prediction pipeline is:
\begin{align}
  \mathcal{L} = \mathcal{L}_{\mathrm{mse}} + \lambda_{\mathrm{lpips}}\mathcal{L}_{\mathrm{lpips}} + \lambda_{\mathrm{ssim}}\mathcal{L}_{\mathrm{ssim}} + \lambda_{\mathbf{v}}\mathcal{L}_{\mathbf{v}} + \lambda_{\boldsymbol{\omega}}\mathcal{L}_{\boldsymbol{\omega}} + \lambda_{\mathbf{l}}\mathcal{L}_{\mathbf{l}} + \lambda_{\mathbf{D}}\mathcal{L}_{\mathbf{D}} + \lambda_{\mathbf{N}}\mathcal{L}_{\mathbf{N}},
  \label{eq:expert_model_loss}
\end{align}
where $\lambda_{\mathrm{lpips}}$, $\lambda_{\mathrm{ssim}}$, $\lambda_{\mathbf{v}}$, $\lambda_{\boldsymbol{\omega}}$, $\lambda_{\mathbf{l}}$, $\lambda_{\mathbf{D}}$ and $\lambda_{\mathbf{N}}$ are the weights for the corresponding loss functions.
We set $\lambda_{\mathrm{lpips}} = 2.0$, $\lambda_{\mathrm{ssim}} = 0.2$, $\lambda_{\mathbf{v}} = 1.0$, $\lambda_{\boldsymbol{\omega}} = 1.0$, $\lambda_{\mathbf{l}} = 1.0$, $\lambda_{\mathbf{D}} = 0.1$ and $\lambda_{\mathbf{N}} = 0.01$ for all experiments.
All weights for the regularization losses are warmed up linearly from $0$ to their final values during the first $2500$ iterations of training.

\section{Implementation Detail}
\label{sec:implementation_detail}

\nbf{Architecture}
We use a modified ViT architecture \cite{dosovitskiy2020image} for our fusion network $\mathcal{F}$.
Specifically, we use $12$ layers of all-to-all self-attention with $16$ heads, each head having a hidden dimension of $96$, and the fully connected layers have a $4\times$ wider hidden channel size.
Since the Plücker coordinates \cite{julius1865new} $\mathbf{P}$ and timestamps $\mathbf{T}$ already provide the 4D position of each pixel, we do not use additional embedding for the positional information.
For the second stage training, where we enable the multi-level spatial-temporal attention module, we copy the weights of the first-stage transformer $L$ times and train them independently.
The $l$-th transformer is responsible for the $l$-th level of spatial-temporal attention, with $1$ classification token for passing information between different levels.
For the MLP decoders $\mathcal{D}$, we use $2$ fully connected layers with a hidden dimension of $256$ for each channel.
Both the transformer modules $\mathcal{F}$ and $\mathcal{G}$ use GELU \cite{hendrycks2016gaussian} as the activation function and layer normalization \cite{ba2016layer} as the normalization function.
We also disable the bias parameters for all the layers.

\nbf{Training \& Inference}
We implement 4DGT in PyTorch framework \cite{paszke2019pytorch}.
We employ FlashAttentionV3 \cite{shah2024flashattention} and the GSplat Rasterizer \cite{ye2025gsplat} for efficient attention and Gaussian optimization respectively.
For optimization, we use the AdamW optimizer \cite{loshchilov2017decoupled} with a learning rate of $5e^{-4}$ and a weight decay of $0.05$.
For the second stage training, the learning rate is set to $1e^{-5}$.
Additionally, we linearly warm-up the learning rate of each stage in the first $2500$ steps and then apply the cosine decaying schedule \cite{loshchilov2016sgdr} for the remaining steps.
During the second strange training, we additionally augment the input and output to the network by varying the aspect ratio and field of view of the images.
Specifically, we randomly sample an aspect ratio from the uniform distribution on $[\frac{1}{3}, \frac{3}{1}]$ and a field of view ratio on the original image on $[30\%, 100\%]$.
We train our reconstruction model $100k$ iterations for the first stage and $30k$ iterations for the second stage, using a total batch size of $64$.
With $64$ Nvidia H100 GPUs, the first stage training takes roughly $9$ days and the second stage training takes roughly $6$ days.
For all other experiments on inference speed, we use a single 80 GB A100 GPU.

\section{Experiments}
\label{sec:experiments}

\nbf{Training Datasets}
\label{sec:datasets}
We use the following real-world monocular videos with high-quality calibrations:
\begin{itemize}[topsep=0pt,leftmargin=*]
    \setlength\itemsep{0em}
    \item Project Aria datasets with closed-loop trajectories: the EgoExo4D \cite{grauman2024ego}, Nymeria \cite{ma2024nymeria}, Hot3D \cite{banerjee2024hot3d} and Aria Everyday Activities (AEA) \cite{lv2024aria}.
    \item Video data with COLMAP \cite{schonberger2016structure} camera parameters: Epic-Fields \cite{tschernezki2023epic,damen2018scaling} and Cop3D \cite{sinha2023common}.
    \item Phone videos with ARKit camera poses: ARKitTrack \cite{zhao2023arkittrack}.
\end{itemize}

\nbf{Evaluation datasets} We use the synthetic rendering provided in ADT \cite{pan2023aria} datasets, which provides metric ground truth depth. To evaluate cross-domain generalization, we use DyCheck \cite{gao2022monocular} (DyC) datasets and the dynamic scene in TUM-SLAM \cite{sturm2012benchmark} (TUM) to evaluate novel view synthesis. We further hold out a test split from EgoExo4D, AEA, and Hot3D, which we refer to as the Aria test set.

\nbf{Metrics}
\label{sec:metrics}
For appearance evaluation, we compare the PSNR and LPIPS \cite{zhang2018unreasonable} metrics on novel view and time rendering results.
For geometry evaluation, we compare against the depth RMSE \cite{yang2024depth} and normal angle error \cite{ye2024stablenormal}.
We additionally provide qualitative comparisons of motion rendered in 2D as optical flow and motion segmentation.
All comparison experiments are conducted on $128$-frame subsequences of the monocular videos, with $64$ frames used as input and the remaining $64$ frames used for testing, with images resized to $504 \times 504$ resolution or a similar pixel number for controlled comparison unless specified otherwise.
For the DyCheck \cite{gao2022monocular} dataset, we additionally compare the rendering results on the provided test view cameras, which show signals on extreme view synthesis.
We provide more details about evaluation implementations in the appendix.

\nbf{Baselines}
\label{sec:baselines}
We consider the following baselines as the most relevant work for evaluation.
\begin{enumerate}[leftmargin=*]
    \setlength\itemsep{0em}
    \item \textbf{L4GM \cite{ren2024l4gm}}: It is the closest prior 4D Gaussian model that generalizes to real-world videos. Different from ours trained using real-world data only, they trained on a synthetic dataset and leveraged additional multi-view diffusion priors from ImageDream \cite{wang2023imagedream}.
    \item \textbf{Static-LRM}: We trained a static scene LRM following \cite{gslrm2024} on the same real world data as our 4DGT. We use 2DGS instead of 3DGS as the representation that shows more similarity to our approach, except that we further model the dynamic content.
    \item \textbf{Expert monocular models}: We compared each individual expert model we used during training in the same setting, including DepthAnythingV2 \cite{yang2024depth} aligned with the metric scale of UniDepth \cite{piccinelli2024unidepth} and normals provided by StableNormal \cite{ye2024stablenormal}. For novel view evaluation, we unproject the image using the nearest depth and normal frame.
    \item \textbf{MonST3R \cite{zhang2024monst3r}}: We compare to the dynamic point based representation \cite{zhang2024monst3r} which highlights the representation difference in using 4DGS. We use ground truth camera poses as input to their model using the official implementation and using PyTorch3D \cite{ravi2020pytorch3d} for normal estimation.
    \item \textbf{Shape of Motion (SoM) \cite{wang2024shape}}: We use SoM to represent the top-tier per-scene optimization method as a reference for best dynamic reconstruction quality. We follow SoM's instructions to manually segment the dynamic part. It requires running expert models as input, including mask, depth, and tracking, which we do not use. We include the preprocessing time for time comparisons.
\end{enumerate}
\camrdy{
    We do not make comparisons with CAT4D \cite{wu2024cat4d}, BulletTimer \cite{liang2024feed} and Pred. 3D Repr. \cite{qi2025predicting} since they provide neither the source code nor the pre-trained models.
}

\begin{figure}[t]
    \centering
    \includegraphics[width=1.0\linewidth]{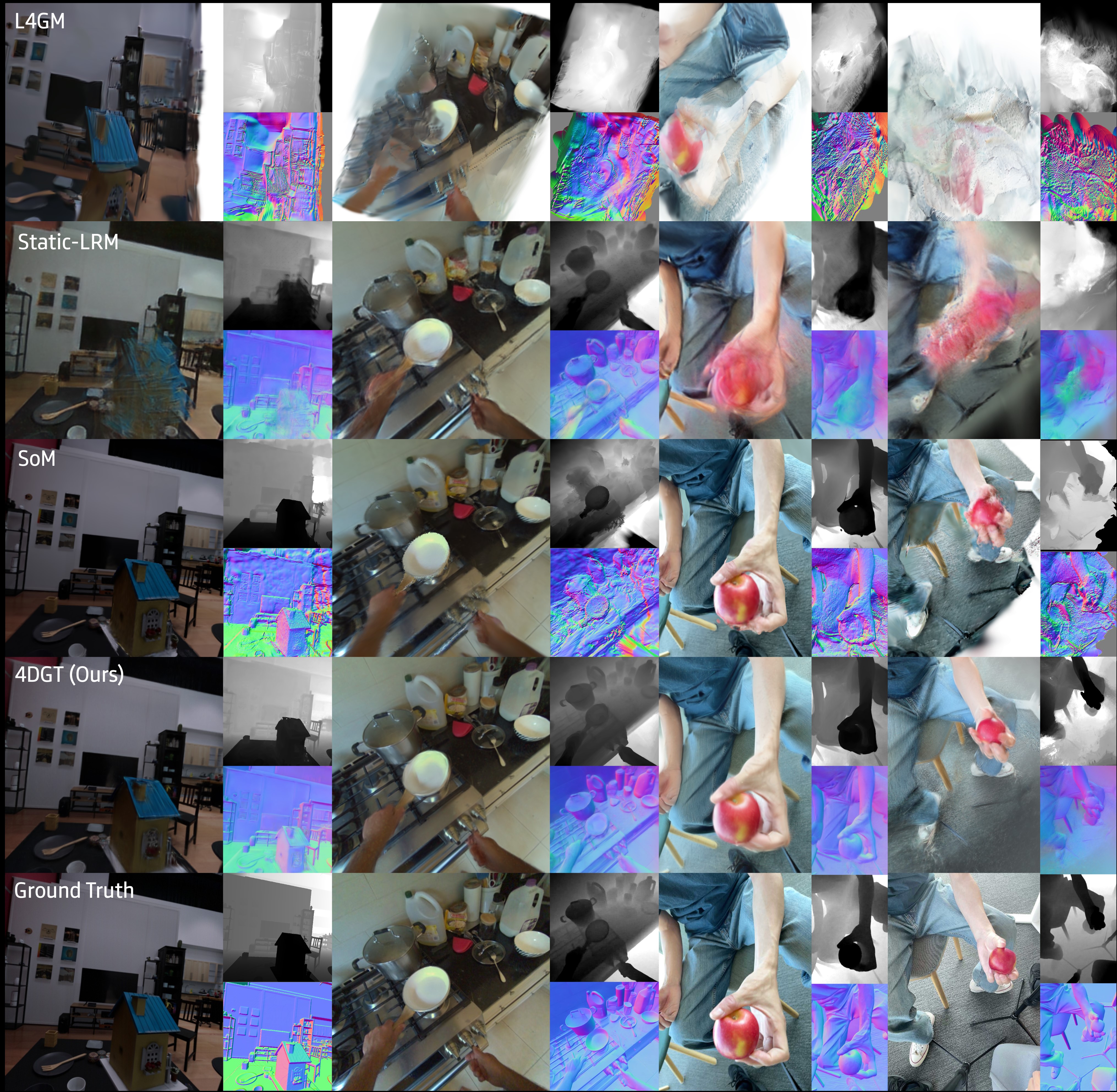}
    \caption{From left-to-right, we show the novel space-time view comparisons on ADT \cite{pan2023aria}, EgoExo4D \cite{grauman2024ego}, DyCheck \cite{gao2022monocular} and the DyCheck test-view (rightmost). We render the depth (upper right) and normal (below right) next to each synthesized novel view. For ground truth depth and normal on EgoExo4D and DyCheck, we use predictions from the expert models from the ground truth image for reference. Please refer to the appendix for more visual comparisons.}
    \label{fig:comparison}
    \vspace{-2em}
\end{figure}

\begin{table}[t]
    \centering
    \caption{Comparisons to baselines. We mark SoM in grey as a reference for optimization-based methods, and rank the other baselines with ours as comparisons in learning based approaches.}

    \begin{adjustbox}{width=\textwidth}
        \setlength{\tabcolsep}{2pt}
        \begin{tabular}{c@{\hskip 5pt}|@{\hskip 5pt}ccccc@{\hskip 5pt}|@{\hskip 5pt}ccccc@{\hskip 5pt}|@{\hskip 5pt}ccc@{\hskip 5pt}|@{\hskip 5pt}c@{\hskip 5pt}|@{\hskip 5pt}c}
            \toprule
            \multirow{2}{*}{Method}                                                     & \multicolumn{5}{c@{\hskip 5pt}|@{\hskip 5pt}}{PSNR (Render) $\uparrow$} & \multicolumn{5}{c@{\hskip 5pt}|@{\hskip 5pt}}{LPIPS (Render) $\downarrow$} & \multicolumn{3}{c@{\hskip 5pt}|@{\hskip 5pt}}{RMSE (Depth) $\downarrow$} & \multicolumn{1}{c@{\hskip 5pt}|@{\hskip 5pt}}{Deg. $\downarrow$} & \multirow{2}{*}{\begin{tabular}[c]{@{}c@{}}Recon.\\Time $\downarrow$\end{tabular}}                                                                                                                                                                                                                                                                                                      \\
                                                                                        & ADT                                                                     & TUM                                                                        & DyC                                                                      & Aria                                                             & \textbf{Avg}                                                                       & ADT                      & TUM                      & DyC                      & Aria                     & \textbf{Avg}                      & ADT                      & TUM                      & \textbf{Avg}                      & ADT                      &                               \\
            \midrule
            \multirow{2}{*}{SoM \cite{wang2024shape}}                                   & \stddev{1.640}                                                          & \stddev{0.266}                                                             & \stddev{2.597}                                                           & \stddev{3.701}                                                   &                                                                                    & \stddev{0.024}           & \stddev{0.069}           & \stddev{0.060}           & \stddev{0.067}           &                                   & \stddev{3.000}           & \stddev{1.354}           & \stddev{1.633}                    & \stddev{3.607}           & \multirow{2}{*}{60000 ms / f} \\
            [-6pt]
                                                                                        & \cellcolor{gray!25}30.30                                                & \cellcolor{gray!25}21.03                                                   & \cellcolor{gray!25}16.49                                                 & \cellcolor{gray!25}26.69                                         & \cellcolor{gray!25}\textbf{23.63}                                                  & \cellcolor{gray!25}0.242 & \cellcolor{gray!25}0.337 & \cellcolor{gray!25}0.392 & \cellcolor{gray!25}0.307 & \cellcolor{gray!25}\textbf{0.320} & \cellcolor{gray!25}4.158 & \cellcolor{gray!25}2.756 & \cellcolor{gray!25}\textbf{3.434} & \cellcolor{gray!25}35.05 &                               \\
            \midrule
            \multirow{2}{*}{L4GM \cite{ren2024l4gm}}                                    & \stddev{2.713}                                                          & \stddev{0.112}                                                             & \stddev{0.629}                                                           & \stddev{1.666}                                                   &                                                                                    & \stddev{0.051}           & \stddev{0.013}           & \stddev{0.062}           & \stddev{0.052}           &                                   & \stddev{0.660}           & \stddev{0.791}           & \stddev{0.386}                    & \stddev{4.181}           &                               \\
            [-6pt]
                                                                                        & 7.348                                                                   & 9.226                                                                      & 8.770                                                                    & 8.617                                                            & \textbf{8.490}                                                                     & 0.688                    & 0.670                    & 0.587                    & 0.698                    & \textbf{0.661}                    & 2.606                    & 1.698                    & \textbf{2.094}                    & 63.51                    & \second{200 ms / f}           \\
            \multirow{2}{*}{MonST3R \cite{zhang2024monst3r,ye2024stablenormal}}         & \stddev{1.651}                                                          & \stddev{1.673}                                                             & \stddev{1.673}                                                           & \stddev{3.355}                                                   &                                                                                    & \stddev{0.011}           & \stddev{0.023}           & \stddev{0.045}           & \stddev{0.135}           &                                   & \stddev{0.842}           & \stddev{0.243}           & \stddev{0.543}                    & \stddev{2.699}           & \multirow{2}{*}{4500 ms / f}  \\
            [-6pt]
                                                                                        & \second{25.13}                                                          & \second{20.61}                                                             & 11.32                                                                    & 19.90                                                            & \second{\textbf{19.24}}                                                            & \second{0.246}           & \best{0.273}             & 0.429                    & 0.323                    & \textbf{0.318}                    & \second{2.111}           & \second{0.653}           & \second{\textbf{1.382}}           & \best{25.00}             &                               \\
            \multirow{2}{*}{Experts \cite{piccinelli2025unidepthv2,ye2024stablenormal}} & \stddev{3.195}                                                          & \stddev{3.117}                                                             & \stddev{1.258}                                                           & \stddev{2.419}                                                   &                                                                                    & \stddev{0.082}           & \stddev{0.073}           & \stddev{0.058}           & \stddev{0.046}           &                                   & \stddev{0.621}           & \stddev{0.074}           & \stddev{0.348}                    & \stddev{0.748}           & \multirow{2}{*}{350 ms / f}   \\
            [-6pt]
                                                                                        & 23.32                                                                   & 18.64                                                                      & \second{11.53}                                                           & 22.32                                                            & \textbf{18.96}                                                                     & 0.299                    & \second{0.318}           & \second{0.423}           & \second{0.236}           & \textbf{0.319}                    & 2.931                    & 0.919                    & \textbf{1.925}                    & 26.26                    &                               \\
            \multirow{2}{*}{\textbf{Ours}}                                              & \stddev{1.508}                                                          & \stddev{0.048}                                                             & \stddev{2.034}                                                           & \stddev{1.591}                                                   &                                                                                    & \stddev{0.021}           & \stddev{0.009}           & \stddev{0.067}           & \stddev{0.019}           &                                   & \stddev{0.463}           & \stddev{0.048}           & \stddev{0.255}                    & \stddev{1.831}           &                               \\
            [-6pt]
                                                                                        & \best{28.31}                                                            & \best{21.02}                                                               & \best{16.12}                                                             & \best{27.36}                                                     & \best{\textbf{23.20}}                                                              & \best{0.243}             & 0.349                    & \best{0.408}             & \best{0.230}             & \best{\textbf{0.308}}             & \best{0.934}             & \best{0.394}             & \best{\textbf{0.664}}             & \second{25.92}           & \best{25 ms / f}              \\
            \bottomrule
        \end{tabular}
    \end{adjustbox}
    \label{tab:comparisons}
\end{table}

\begin{table}[t]
    \centering
    \caption{Ablation study on our method components using ADT and DyCheck (DyC).}

    \begin{subtable}{0.505\textwidth}
        \centering
        \caption{Ablation on dynamic Gaussian in stage one training.}
        \begin{adjustbox}{width=\textwidth}
            \setlength{\tabcolsep}{2pt}
            \begin{tabular}{c@{\hskip 5pt}|@{\hskip 5pt}ccc@{\hskip 5pt}|@{\hskip 5pt}ccc@{\hskip 5pt}|@{\hskip 5pt}c@{\hskip 5pt}|@{\hskip 5pt}c}
                \toprule
                \multirow{2}{*}{Method}                                                                                              & \multicolumn{3}{c@{\hskip 5pt}|@{\hskip 5pt}}{PSNR$\uparrow$} & \multicolumn{3}{c@{\hskip 5pt}|@{\hskip 5pt}}{LPIPS$\downarrow$} & \multicolumn{1}{c@{\hskip 5pt}|@{\hskip 5pt}}{RMSE} & \multicolumn{1}{c}{Deg.}                                                                                \\
                                                                                                                                     & ADT                                                           & DyC                                                              & \textbf{Avg}                                        & ADT                      & DyC            & \textbf{Avg}            & ADT$\downarrow$ & ADT$\downarrow$ \\
                \midrule
                Naive                                                                                                                & 15.49                                                         & 13.95                                                            & \textbf{14.72}                                      & 0.612                    & 0.517          & \textbf{0.564}          & 1.156           & 42.25           \\
                + EgoExo4D \cite{grauman2024ego}                                                                                     & \second{22.72}                                                & \second{15.27}                                                   & \second{\textbf{19.00}}                             & \second{0.229}           & \second{0.385} & \second{\textbf{0.307}} & 1.278           & 41.11           \\
                Static LRM \cite{gslrm2024}                                                                                          & 19.29                                                         & 14.21                                                            & \textbf{16.75}                                      & 0.399                    & 0.463          & \textbf{0.431}          & \second{0.830}  & \second{31.96}  \\
                Per-frame \cite{ren2024l4gm}                                                                                         & 10.07                                                         & 12.10                                                            & \textbf{11.08}                                      & 0.748                    & 0.714          & \textbf{0.731}          & 3.117           & 61.77           \\
                \textbf{+ $\boldsymbol{\mathcal{L}}_{\boldsymbol{\mathbf{N},\mathbf{D},\mathbf{l},\boldsymbol{\omega},\mathbf{v}}}$} & \best{26.45}                                                  & \best{15.86}                                                     & \best{\textbf{21.15}}                               & \best{0.170}             & \best{0.399}   & \best{\textbf{0.284}}   & \best{0.773}    & \best{21.59}    \\
                \bottomrule
            \end{tabular}
        \end{adjustbox}
        \label{tab:ablations_feedforward}
    \end{subtable}
    \hfill
    \begin{subtable}{0.486\textwidth}
        \centering
        \caption{Ablation on stage two training.}
        \begin{adjustbox}{width=\textwidth}
            \setlength{\tabcolsep}{2pt}
            \begin{tabular}{c@{\hskip 5pt}|@{\hskip 5pt}ccc@{\hskip 5pt}|@{\hskip 5pt}ccc@{\hskip 5pt}|@{\hskip 5pt}c@{\hskip 5pt}|@{\hskip 5pt}c}
                \toprule
                \multirow{2}{*}{Method} & \multicolumn{3}{c@{\hskip 5pt}|@{\hskip 5pt}}{PSNR$\uparrow$} & \multicolumn{3}{c@{\hskip 5pt}|@{\hskip 5pt}}{LPIPS$\downarrow$} & \multicolumn{1}{c@{\hskip 5pt}|@{\hskip 5pt}}{RMSE} & \multicolumn{1}{c}{Deg.}                                                                                \\
                                        & ADT                                                           & DyC                                                              & \textbf{Avg}                                        & ADT                      & DyC            & \textbf{Avg}            & ADT$\downarrow$ & ADT$\downarrow$ \\
                \midrule
                Naive                   & \multicolumn{8}{c}{\cellcolor{gray!25}Out of Memory}                                                                                                                                                                                                                                             \\
                Random.                 & 25.79                                                         & 14.97                                                            & \textbf{20.38}                                      & 0.333                    & 0.480          & \textbf{0.406}          & \second{0.750}  & 25.84           \\
                + D\&P                  & \best{28.79}                                                  & \second{15.52}                                                   & \second{\textbf{22.16}}                             & \second{0.242}           & \second{0.434} & \second{\textbf{0.338}} & \best{0.722}    & 25.95           \\
                + Multi-level           & \second{28.71}                                                & 15.34                                                            & \textbf{22.03}                                      & \second{0.242}           & 0.439          & \textbf{0.341}          & 0.783           & 27.30           \\
                \textbf{+ Mix. (Ours)}  & 28.31                                                         & \best{16.12}                                                     & \best{\textbf{22.22}}                               & 0.243                    & \best{0.408}   & \best{\textbf{0.326}}   & 0.934           & \second{25.92}  \\
                \bottomrule
            \end{tabular}
        \end{adjustbox}
        \label{tab:ablations_density_control}
    \end{subtable}

    \begin{subtable}{0.620\textwidth}
        \centering
        \caption{Evaluation on the dynamic foreground.}
        \begin{adjustbox}{width=\textwidth}
            \setlength{\tabcolsep}{2pt}
            \begin{tabular}{c@{\hskip 5pt}|@{\hskip 5pt}ccc@{\hskip 5pt}|@{\hskip 5pt}ccc@{\hskip 5pt}|@{\hskip 5pt}c@{\hskip 5pt}|@{\hskip 5pt}c}
                \toprule
                \multirow{2}{*}{Method}              & \multicolumn{3}{c@{\hskip 5pt}|@{\hskip 5pt}}{PSNR$\uparrow$} & \multicolumn{3}{c@{\hskip 5pt}|@{\hskip 5pt}}{LPIPS$\downarrow$} & \multicolumn{1}{c@{\hskip 5pt}|@{\hskip 5pt}}{RMSE} & \multicolumn{1}{c}{Deg.}                                                                            \\
                                                     & ADT                                                           & DyC                                                              & \textbf{Avg}                                        & ADT                      & DyC          & \textbf{Avg}          & ADT$\downarrow$ & ADT$\downarrow$ \\
                \midrule
                Static LRM \cite{gslrm2024} (masked) & 17.30                                                         & 13.56                                                            & \textbf{16.76}                                      & 0.059                    & 0.220        & \textbf{0.102}        & 0.613           & 49.35           \\
                \textbf{Ours (masked)}               & \best{27.29}                                                  & \best{14.93}                                                     & \best{\textbf{22.86}}                               & \best{0.030}             & \best{0.195} & \best{\textbf{0.075}} & \best{0.388}    & \best{33.32}    \\
                \bottomrule
            \end{tabular}
        \end{adjustbox}
        \label{tab:ablations_masked}
    \end{subtable}
    \hfill
    \begin{subtable}{0.365\textwidth}
        \centering
        \caption{\camrdy{Motion segmentation results.}}
        \vspace{8pt}
        \begin{adjustbox}{width=\textwidth}
            \setlength{\tabcolsep}{2pt}
            \camrdy{
                \begin{tabular}{lccc}
                    \toprule
                    Method          & w/o $\mathcal{L}_{\mathbf{v},\boldsymbol{\omega},\mathbf{l}}$ & MegaSaM~\cite{li2024megasam} & \textbf{Ours}            \\
                    \midrule
                    mIoU $\times 100$ $\uparrow$ & $9.4^{\pm4.1}$                                                & $77.4^{\pm4.0}$              & $\mathbf{81.2^{\pm1.8}}$ \\
                    \bottomrule
                \end{tabular}
            }
        \end{adjustbox}
        \label{tab:ablations_motion_mask}
    \end{subtable}

\end{table}

\begin{table}[t]
    \centering
    \caption{\camrdy{Comparison with a more comprehensive upper-bound on the ADT dataset~\cite{pan2023aria}.}}
    \begin{adjustbox}{width=0.75\textwidth}
        \setlength{\tabcolsep}{2pt}
        \begin{tabular}{lccccc}
            \toprule
            Method                                & PSNR $\uparrow$              & LPIPS $\downarrow$            & RMSE $\downarrow$             & Degree $\downarrow$         & Recon. Time $\downarrow$ \\
            \midrule
            SoM~\cite{wang2024shape}              & \second{30.30$^{\pm1.64}$}   & \second{0.242$^{\pm0.024}$}   & 4.16$^{\pm3.00}$              & 34.07$^{\pm3.61}$           & 60,000 ms/f          \\
            SoM*~\cite{wang2024shape,kerbl20233d} & 28.40$^{\pm2.60}$            & 0.281$^{\pm0.027}$            & 4.18$^{\pm2.98}$               & \best{18.46$^{\pm1.76}$}    & 60,000 ms/f           \\
            \textbf{Ours}                         & 28.31$^{\pm1.51}$            & 0.243$^{\pm0.021}$            & \second{0.93$^{\pm0.46}$}      & 25.94$^{\pm1.84}$           & \best{25 ms/f}         \\
            \textbf{Ours$_{\text{tune10s}}$}      & \best{31.98$^{\pm1.01}$}     & \best{0.220$^{\pm0.012}$}     & \best{0.85$^{\pm0.45}$}        & \second{19.25$^{\pm2.65}$}  & \second{25 + 150 ms/f} \\
            \bottomrule
        \end{tabular}
    \end{adjustbox}
    \label{tab:init_optimization}
\end{table}

\nbf{Comparisons to baselines} Table~\ref{tab:comparisons} and Figure~\ref{fig:comparison} present our comparisons to baselines. Compared to L4GM, which also predicts dynamic Gaussians from a trained transformer, 4DGT shows much better generalization across real scenes. We found that a static-LRM can provide a strong baseline for static scenes but will fail when dynamic motion is present, while 4DGT can do well in both. It is further validated in Table~\ref{tab:ablations_masked} on dynamic regions. The geometry predicted from 4DGT is more consistent with the world coordinate compared to expert models when evaluated in metric scale. Compared to the optimization-based method SoM, 4DGT can offer on-par quality in view synthesis as well as geometry prediction while being 3 orders of magnitude faster in runtime, which makes it more favorable to process long-time videos in practice.

\nbf{Ablation study in stage one training} In Table~\ref{tab:ablations_feedforward} for stage one training, we start from a \emph{Naive} training baseline at coarse resolution using the image rendering losses in Eq.~\ref{eq:imag_rendering_loss} trained only using the AEA dataset with only 7 hours of data. After further scaling to using the EgoExo4D dataset (~300 hours), we find that increasing the scale of the dataset can significantly improve the performance. We also compare to a static LRM \cite{gslrm2024} and per-frame LRM \cite{ren2024l4gm} counterpart in the same setting. We can already see the benefits over the baselines at this stage by a large margin. We further include the full training loss in Eq.~\ref{eq:regularization_loss} and Eq.~\ref{eq:expert_model_loss}, and we can see significant improvements in the quality across all metrics. \camrdy{The regularization terms $\boldsymbol{\mathcal{L}}_{\boldsymbol{\mathbf{N},\mathbf{D},\mathbf{l},\boldsymbol{\omega},\mathbf{v}}}$ can be be categorized into two groups: (1) $\mathcal{L}_{\mathbf{v},\boldsymbol{\omega},\mathbf{l}}$ regularizes the motion of the dynamic Gaussian predictions. Without this term, although the quantitative results are not greatly affected, the model would fall into the trivial local minima of making every Gaussian transient and dynamic, not correctly modeling the scene's static or slow-moving parts. This would result in a purely-white motion mask. In \cref{tab:ablations_motion_mask}, we evaluate the quality of the motion mask on the ADT dataset \cite{pan2023aria} of our method without this term. Thanks to the explicit modeling of the motion parameters in our representation, our model can produce comparable motion segmentation against MegaSaM \cite{li2024megasam}, which has explicit flow supervision, while being 200$\times$ faster (1.5s v.s. 300s). (2) $\mathcal{L}_{\mathbf{N},\mathbf{D}}$ provides expert guidance for the geometry of the dynamic Gaussian prediction. These terms can greatly improve the quality of the reconstructed geometry.
}
Please refer to the appendix for visual comparisons.

\nbf{Ablation study in stage two training} Table.~\ref {tab:ablations_density_control} shows the ablation study of key design in stage two training. Starting from a \emph{Naive} model without density control to prune and densify (\emph{D\&P}) Gaussians and the \emph{multi-level} attention proposed in \ref{sec:density_control}, naively scaling up resolution and temporal samples will run out of memory in training. Compared with variants using a \emph{Random} sampled Gaussian from predicted patches and further densifying, using the proposed D\&P strategy to decode sparse activated Gaussian will lead to better results while being efficient in training. Adding the proposed multi-level attention can further speed up training with only a minor sacrifice in quality, but can speed up training two times faster. Finally, we mixed all the proposed datasets in training at stage two. Compared to model training only using EgoExo4D, mixing datasets improves generalization across domains.

\nbf{Initializing optimization-based methods with 4DGT}
\Cref{tab:init_optimization} presents a quantitative comparison on the ADT dataset~\cite{pan2023aria} with stronger baselines and 4DGT-initialized optimization-based method. \textit{SoM-2DGS-Geometry} augments SoM~\cite{wang2024shape} with 2DGS~\cite{kerbl20233d} and employs the same normal regularization as ours to provide a stronger geometry upper-bound. This results in improved normal quality, but our feed-forward 4DGT prediction method still achieves comparable—or superior—results while being orders of magnitude faster. After finetuning the feed-forward prediction for only 10 seconds (100 iterations, 150\,ms per frame, denoted as \emph{Ours$_{\text{tune10s}}$}), performance further improves beyond all optimization-based baselines. While SoM and its 2DGS-augmented variant require 30{,}000 optimization steps per scene (and a sophisticated tracking expert like TAPIR~\cite{doersch2023tapir}), our finetuned feed-forward prediction achieves a $350\times$ speedup, and pure feed-forward performance (\emph{Ours}) is $2,400\times$ faster. More finetuning further improves reconstruction quality.

\section{Conclusion}
\label{sec:conclusion}

We introduced 4DGT, a novel dynamic scene reconstruction method that predicts 4DGS from an input posed video frame in a feed-forward manner. The representation power of 4DGT enables it to handle general dynamic scenes using 4DGS with varying lifespans, and support it to handle complex dynamics in long videos. Different from prior work that heavily depends on multi-view supervision from synthetic datasets, 4DGT is trained only using real-world monocular videos. We demonstrate that 4DGT can generalize well to videos recorded from similar devices, and the ability of generalization can improve when mixing datasets for training in scale.

\nbf{Limitations and future work} We do not claim 4DGT can generalize to all videos in the wild. We assume the availability of a reliable calibration to train 4DGT and deploy it for inference. For this requirement, the training datasets have been limited to data sources from a few egocentric devices and phone captures. We observe that the quality may degrade in videos recorded by an unseen type of device due to the inaccurate metric scale calibrations. We believe this can be significantly improved by further scaling up the method using more diverse datasets recorded by different devices with curated calibrations. Similar to most monocular reconstruction methods, we still observe significant artifacts when viewing Gaussians from extreme view angles, departing far from the input trajectory. Future directions can propose better representations to address it or learn to distill more priors from multi-view expert models, such as generative video models.

\clearpage
\begin{ack}
    We thank Aljaz Bozic for the insightful discussions.
    We thank the Project Aria team for their open-source and dataset contributions.
    This work was also partially supported by NSFC (No. U24B20154).
\end{ack}

{
    \small
    \bibliographystyle{abbrvnat}
    \bibliography{main}
}

\clearpage

\begingroup
\tableofcontents
\endgroup
\clearpage

\appendix

\section{Additional Details}

\subsection{Additional Details on the Multi-Level Attention Module}

As mentioned in the Method section of the main paper, aside from the number of Gaussians, another efficiency-limiting factor is the number of tokens in the large number of high-resolution images.
In the second stage of training, we increase the spatial and temporal resolution by a factor of $R_{s}$ and $R_{t}$, respectively.
The number of patches participating in the self-attention module $\mathcal{F}$ increases by a factor of $R_{s}^2 \cdot R_{t}$, which will slow down optimization and inference significantly.
To mitigate this, we propose a temporal level-of-detail attention mechanism to reduce the computational cost.
Specifically, noticing that the computational complexity of the self-attention module is $O(n^2)$ (simplified from $O(n^2+n)$) where $n$ is the number of tokens \cite{vaswani2017attention},
We propose to divide the $N$ input frames into $M$ equal trunks in the highest level.
This division limits the attention mechanism in the temporal dimension, but reduces the computation of calculating $n$ total tokens to $O(\frac{n^2}{M})$.
To balance spatial-temporal samples, we construct a temporal level-of-detail structure by alternating the temporal range and spatial resolution, achieving a much smaller overhead while maintaining the ability to handle long temporal windows.
For each level $l$, we reduce the spatial resolution by a factor of $2^l$ and increase temporal samples by 2.
This results in a computational complexity of:
\begin{align}
    O(\frac{n^2}{M} + \cdots + \frac{n^2}{{M} \cdot {2^{L-1}}}) = O(\frac{n^2}{M \cdot 2^{L-1}} \cdot \sum_{l=0}^{L-1} 2^l) = O(n^2 \cdot \frac{2^L-1}{M \cdot 2^{L - 1}}) \approx O(\frac{2n^2}{M}).
\end{align}
Empirically, we use level $L=3$ and $M=4$, which leads to an approximately $2$ times reduction in the computational cost.

\subsection{Additional Details on the Densification \& Pruning of the Dynamic Gaussian}

\begin{figure}[t]
    \centering
    \label{fig:opacity}
    \includegraphics[width=1.0\textwidth]{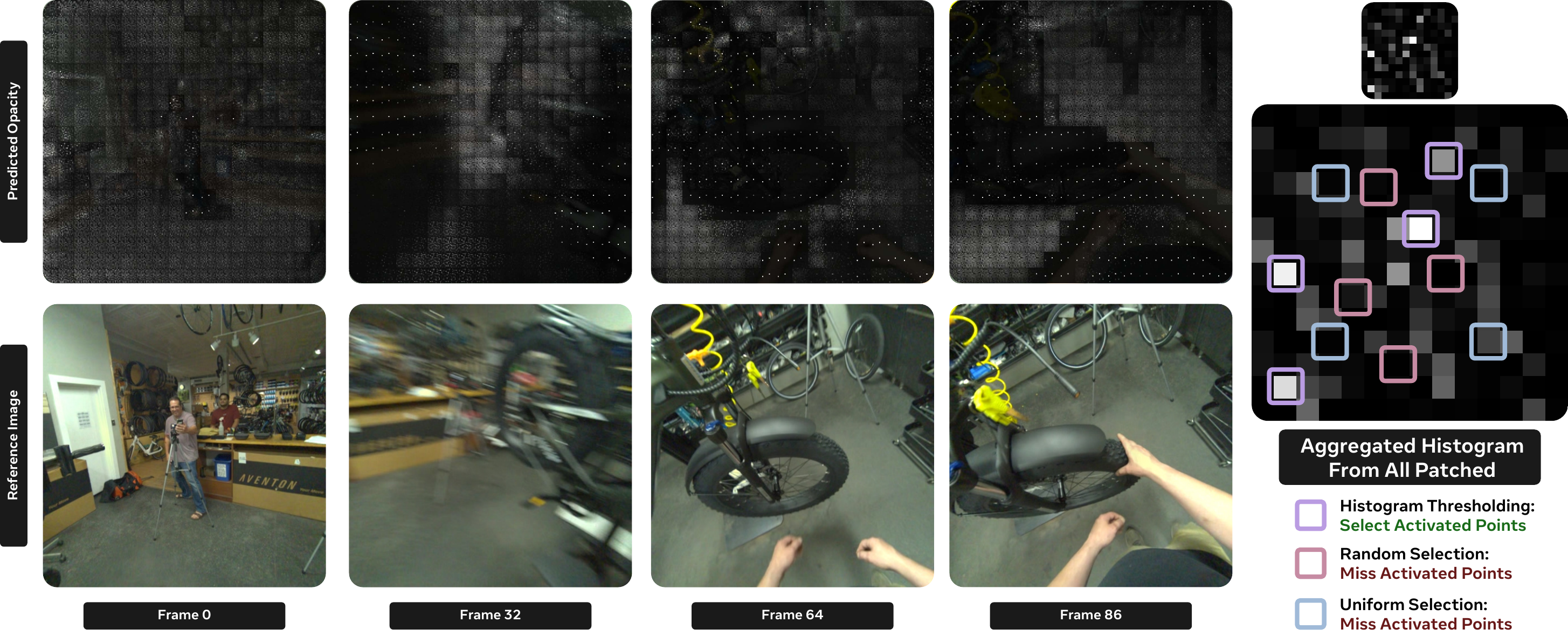}
    \caption{The predicted opacity map ($\in R^{N \times H \times W}$) of the pixel-aligned dynamic Gaussians from 4DGT and the computed histogram ($\in R^{p \times p}$) of the activation distribution. The right section shows the difference between histogram thresholding (Ours) and other filtering methods (randomly or uniformly selecting the Gaussians to keep) for reducing the number of Gaussians.}
\end{figure}

In \cref{fig:opacity}, we visualize the predicted opacity map of the pixel-aligned dynamic Gaussians.
It shows clear patterns of the activation of the pixel inside each patch, especially for the dynamic regions.
Notably, randomly or uniformly selecting the Gaussians to keep will lead to a significant number of active Gaussians being pruned, effectively removing the ability of the model to model the dynamic parts, while our histogram thresholding scheme can effectively keep the Gaussians that are contributing.
These strategies blend the densification and pruning strategies of Gaussian representations \cite{kerbl20233d} and the multi-stage training strategy of ViT models \cite{wang2025cut3r}, effectively introducing a density control scheme for the feed-forward prediction pipeline.

\subsection{Additional Details on Datasets and Baselines}

For each dataset used in training \cite{lv2024aria,grauman2024ego,banerjee2024hot3d,zhao2023arkittrack,ma2024nymeria,sinha2023common,tschernezki2023epic}, we select $99.15\%$ of the sequences as the training set and hold out the rest.
For the datasets used in evaluation:
\begin{itemize}
    \item \textbf{ADT} \cite{pan2023aria}: We select 4 subsequences for validating the reconstruction performance:
          \begin{itemize}
              \item \textit{Apartment\_release\_multiuser\_cook\_seq141\_M1292}
              \item \textit{Apartment\_release\_multiskeleton\_party\_seq114\_M1292}
              \item \textit{Apartment\_release\_meal\_skeleton\_seq135\_M1292}
              \item \textit{Apartment\_release\_work\_skeleton\_seq137\_M1292}
          \end{itemize}
    \item \textbf{DyCheck} \cite{gao2022monocular}: We use all 6 sequences with 3 views, and follow \cite{wang2024shape,gao2022monocular} to apply the covisibility mask before computing metrics on novel views:
          \begin{itemize}
              \item \textit{apple}, \textit{block}, \textit{space-out}, \textit{spin}, \textit{paper-windmill}, \textit{teddy}
          \end{itemize}
    \item \textbf{TUM} \cite{sturm2012benchmark}: We seclet 3 subsequences for evaluation:
          \begin{itemize}
              \item \textit{rgbd\_dataset\_freiburg2\_desk\_with\_person}
              \item \textit{rgbd\_dataset\_freiburg3\_walking\_halfsphere}
              \item \textit{rgbd\_dataset\_freiburg3\_sitting\_halfsphere}
          \end{itemize}
    \item \textbf{EgoExo4D} \cite{grauman2024ego}: We select 3 subsequences from the hold-out sequences:
          \begin{itemize}
              \item \textit{cmu\_bike01\_2}, \textit{sfu\_cooking015\_2}, \textit{uniandes\_bouldering\_003\_10}
          \end{itemize}
    \item \textbf{Nymeria} \cite{ma2024nymeria}: We select 2 sequences from the hold-out set:
          \begin{itemize}
              \item \textit{20230607\_s0\_james\_johnson\_act1\_7xwm28}
              \item \textit{20230612\_s1\_christina\_jones\_act0\_u2r0z8}
          \end{itemize}
    \item \textbf{AEA} \cite{lv2024aria}: We select the \textit{loc5\_script5\_seq7\_rec1} sequence from the hold-out set.
    \item \textbf{Hot3D} \cite{banerjee2024hot3d}: We select the \textit{P0020\_ff537251} sequence from the hold-out set.
\end{itemize}
The testing sequences from EgoExo4D, AEA, and Hot3D are denoted as \textit{Aria} in all comparisons.

Note that we do not make comparisons with CAT4D \cite{wu2024cat4d}, BulletTimer \cite{liang2024feed} since they provide neither the source code nor the pre-trained models as of writing.

\camrdy{

    \subsection{Additional Details on the Number of Gaussians}

    The number of dynamic Gaussians predicted by the first stage is pixel-aligned, and can be computed as (derived from \cref{eq:gs_prediction}):
    \begin{equation}
        N_{\mathbf{g}} = N \times H \times W.
    \end{equation}
    For a resolution of $252 \times 252$ and $16$ images (half spatial resolution and $1/4\times$ temporal resolution of the second stage), this results in $508{,}032$ (0.5M) Gaussians.

    In the second stage, the resolution is increased to $504 \times 504$ and $64$ images. With the proposed patch-based pruning strategy, the number of Gaussians can be computed as (derived from \cref{eq:gs_prediction}):
    \begin{equation}
        N_{\mathbf{g}} = N \times H \times W \times \frac{S}{p^2},
    \end{equation}
    which results in a total of $829{,}440$ Gaussians.

    Finally, the proposed multi-level spatial attention mechanism introduces two additional downsampled outputs with $1/4\times$ and $1/16\times$ the spatial resolution, respectively. This leads to a final Gaussian count of:
    \begin{equation}
        N_{\mathbf{g}} = 829{,}440 \times \left(1 + \frac{1}{4} + \frac{1}{16}\right) = 1{,}088{,}640~\text{(1M)}.
    \end{equation}
    for the second stage.

    Thanks to our proposed selective activation pruning strategy, the number of Gaussians only increases by $1\times$ while the space-time resolution increases $15\times$.

}

\section{Additional Results}

\subsection{Qualitative Results of the Ablation Study}

\begin{figure}[ht]
    \centering
    \includegraphics[width=1.0\textwidth]{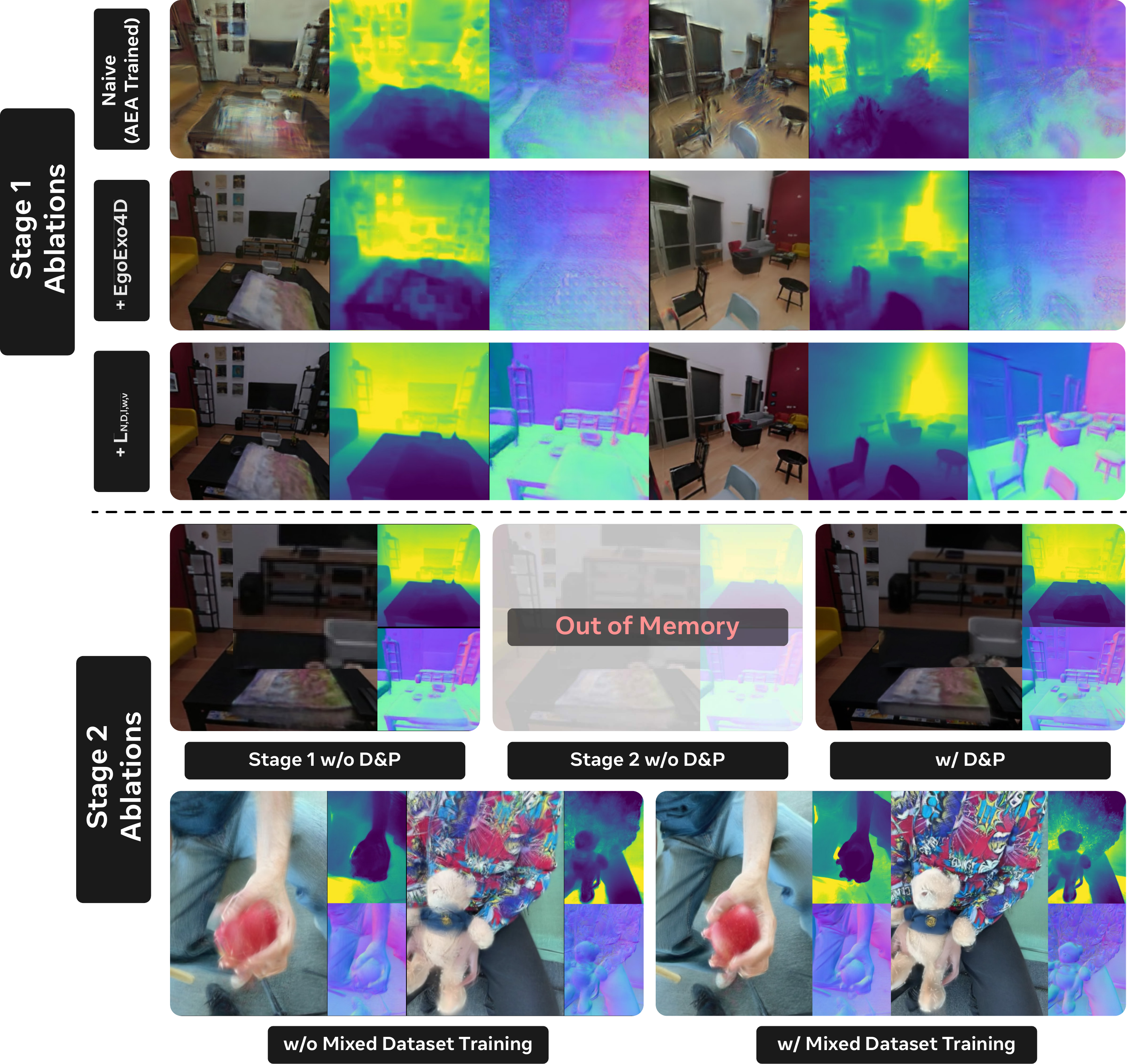}
    \caption{Ablation study on proposed components.}
    \label{fig:ablations}
\end{figure}

In \cref{fig:ablations}, we show the ablation study results for the first and second stage training, respectively.
As shown in the table and figure, our proposed loss and representation effectively model the dynamic regions and improve the reconstruction quality.
Moreover, the proposed density control scheme effectively regularized the number of Gaussians with increased input count and input resolution, greatly improving details and avoiding using too much memory.
Adding larger-scale datasets \cite{grauman2024ego,zhao2023arkittrack} helps generalization for both in-domain and out-of-domain datasets.

\camrdy{
    \subsection{Results and Discussion on Pruning Pattern Selection}

    In \cref{eq:histogram_calculation}, after computing $\mathbf{H}$ once following the first-stage training, the top $S$ entries are selected to define a shared pruning pattern for the second-stage training. This approach effectively shares the same pattern across all patches.

    The motivation for this design is twofold:
    \begin{itemize}
        \item \textbf{Empirical activation consistency:} Across the $p^2$ pixels within each patch, the model consistently favors similar pixels for activation and subsequent use in Gaussian rendering across all patches. This aggregation leads to a clear shared activation pattern, as visualized in \cref{fig:opacity} (right), where the predicted opacity maps exhibit this pattern consistently. Quantitative results (\cref{tab:pruning_ablation}) show that the shared pruning pattern achieves on-par or better performance compared to recalculating the pattern for each patch on-the-fly.
        \item \textbf{Implementation efficiency:} Using a shared pruning pattern enables efficient implementation by discarding unused rows in the weight matrix of the final fully-connected layer of the decoder heads. This is considerably less resource-intensive regarding both memory usage and computation time, compared to dynamically sorting the opacity values for every patch during runtime.
    \end{itemize}

    \begin{table}[ht]
        \centering
        \caption{\camrdy{\textbf{Comparison of pruning strategies on the ADT~\cite{pan2023aria} dataset.} ``On-the-fly" computes a unique pruning pattern for each patch. ``Shared'' (ours) uses a single pattern for all patches.}}
        \label{tab:pruning_ablation}
        \camrdy{
            \begin{tabular}{lccccc}
                \toprule
                Method                 & PSNR$\uparrow$   & LPIPS$\downarrow$ & RMSE$\downarrow$ & Degree$\downarrow$ & Speed Overhead \\
                \midrule
                On-the-fly             & $28.36$          & $\mathbf{0.241}$  & $0.78$           & $25.95$            & Yes            \\
                \textbf{Shared (Ours)} & $\mathbf{28.79}$ & $0.242$           & $\mathbf{0.72}$  & $\mathbf{25.84}$   & \textbf{No}    \\
                \bottomrule
            \end{tabular}
        }
    \end{table}

}

\subsection{Supplementary Videos}

We attach additional video results in the supplementary video material.
The supplementary video is structured as follows:
\begin{itemize}
    \item \texttt{00:00:00-00:00:15}: Brief introduction to the input \& output setting and goal of the paper.
    \item \texttt{00:00:15-00:00:45}: Reconstruction and novel view rendering results for the depth, normal, optical flow, dynamic mask, and appearance inferred in a rolling window fashion over a long video.
    \item \texttt{00:00:45-00:01:05}: More qualitative video results from other datasets.
    \item \texttt{00:01:05-00:01:35}: Comparison with baseline methods StaticLRM \cite{gslrm2024}, L4GM \cite{ren2024l4gm} and Shape-of-Motion \cite{wang2024shape}.
    \item \texttt{00:01:35-00:02:00}: Ablation study of the proposed components.
\end{itemize}

\camrdy{

    \section{Additional Disucssions}

    \subsection{Additional Discussions on Expert Models}

    Aside from the expert normal and depth guidance, one natural way to improve the outputs' temporal consistency is to incorporate a flow expert's guidance \cite{teed2020raft}.
    It seems an optical flow expert like RAFT \cite{teed2020raft} can easily be plugged into our pipelin,e but it's non-trivial in practice. There are two reasons we did not adopt such a flow model for guidance:
    \begin{itemize}
        \item In our preliminary experiments, we discovered that the estimated optical flow exhibits strong inconsistency, often reaching more than 10 pixels in cycle consistency errors. This in turn makes the training of the 4DGT model unstable and leads to NaN values in the prediction.
        \item A tracking expert model like TAPIR \cite{doersch2023tapir} would produce much more consistent results for guiding the prediction of dynamic Gaussians, as shown by Shape-of-Motion \cite{wang2024shape}. However, the computation of such dense all-to-all tracking is extremely time-consuming (a few hours for a 128-frame clip), making it impractical for our large-scale training setup (1000 hours of video data).
    \end{itemize}
    Due to these reasons, we leave the addition of the flow expert model's guidance to improve the temporal consistency to future work.

    \subsection{Additional Discussions on the Choice of the Dynamic Gaussian Representation}

    The main purpose for our choice of dynamic Gaussian representation is to enable seamless integration to a feed-forward prediction pipeline, which can be used for self-supervised training on general dynamic videos.
    Compared to per-frame 3DGS, static 3DGS, flow vector field or decomposed motion bases, we found the explicit modeling of the motion terms of dynamic Gaussians (adapted from FTGS \cite{yang2023gs4d}, originally proposed in 4DGS \cite{kerbl20233d} (L124, L129)) to be a better fit for this purpose.
    \begin{itemize}
        \item Compared to a per-frame 3DGS \cite{yang2023gs4d} representation (denoted as the *per-frame* variant in the ablation studies of the paper), our representation enables the integration of space-time information as a 4D Gaussian with a non-zero life-span, automatically encodes information across multiple frames, making it possible to train the 4DGT model in a self-supervised manner on monocular videos. In the most extreme case with infinite life-span, the representation is reduced to a purely static 3DGS (denoted as the *Static-LRM* baseline in the paper) and would only work on static scenes.
              Compared to per-frame 3DGS and static 3DGS, our representation can freely encode the different levels of motion speed (from $0$ to $\infty$) of the dynamic scene. Comparison against the *StaticLRM* baseline and *per-frame 3DGS* variant can be found in Table 2(a) of the main paper.
        \item Compared to a 3D flow vector field representation, like DynamicGaussians \cite{yang2023gs4d}, our representation can be easily integrated into the pixel-aligned feed-forward prediction pipeline for patch-based vision transformers. However, their flow vector field, which is typically encoded by an MLP, would be much harder to predict in a feed-forward manner.
              Similar problems exist for the rigid motion representation used in Shape-of-Motion \cite{wang2024shape}, since it's extremely ill-posed to accurately predict their motion bases and coefficients without complicated initialization.
              In comparison, our representation does not require such careful initialization. In practice, we simply set all $\mathbf{v}, \boldsymbol{\omega}$ to zero, $\mathbf{t}$ to the timestamp of the corresponding frame, and $\mathbf{l}$ to a large value (50s).
        \item Compared to other implicit network-based methods like NeRF \cite{mildenhall2020nerf} (as used in Pred. 3D Repr. \cite{qi2025predicting}), a Gaussian-based representation would enable much more efficient rendering and training.
    \end{itemize}

    \subsection{Additional Discussions on Metric Scale Cameras}
    We empirically find the model works best when trained and inferred with metric-scale cameras due to the ambiguity in depth-scale estimation. Notably, the model doesn't rely on fully accurate scaling to perform well, as shown by experiments on the COP3D and EPIC-FIELDS datasets \cite{sinha2023common,tschernezki2023epic}, showing the ability to handle slight deviation from metric-scale calibrations. By introducing such non-metric datasets in training, we force the model to reason from the relative relation of the input cameras and the input images. However, the model would fail to predict coherent results when there exists an order-of-magnitude scale error.

    \subsection{Additional Discussions on Limitations}
    Due to the ill-posed nature of monocular reconstruction (e.g., limited viewpoint coverage and low frame rates), our method, like other monocular approaches, can still exhibit some blurriness and artifacts, especially during sudden or very fast movements and when visualizing reconstructions from challenging viewpoints. These issues are largely inherent to current monocular reconstruction paradigms. Notably, however, as also pointed out by reviewers, our approach already surpasses prior state-of-the-art baselines in terms of sharpness and artifact reduction, and demonstrates results comparable to optimization-based methods. Further improvements, such as scaling up the training datasets and introducing additional expert or supervisory signals to the 4DGT model, are promising directions for alleviating these remaining limitations.

}

\section{Social Impact}

As an early-stage research on 4D reconstruction, we do not foresee any immediate social impact from this work.
However, it's worth noting that such a feed-forward pipeline could be used to synthesize more convincing fake videos by introducing novel views.

\end{document}